# Review of the Assurance of Machine Learning for use in Autonomous Systems (AMLAS) Methodology for Application in Healthcare

Shakir Laher[1,4] *, Carla Brackstone[2], Sara Reis[3], An Nguyen[3], Sean White[1], Ibrahim Habli[4]

* Correspondence: shakir.laher1@nhs.net
　　　　　　　　　sl2318@york.ac.uk

1. NHS Digital
2. Kheiron Medical Technologies
3. Health Navigator
4. University of York

## Abstract

In recent years, the number of machine learning (ML) technologies gaining regulatory approval for healthcare has increased significantly allowing them to be placed on the market. However, the regulatory frameworks applied to them were originally devised for traditional software, which has largely rule-based behaviour, compared to the data-driven and learnt behaviour of ML. As the frameworks are in the process of reformation, there is a need to proactively assure the safety of ML to prevent patient safety being compromised. The Assurance of Machine Learning for use in Autonomous Systems (AMLAS) methodology was developed by the Assuring Autonomy International Programme based on well-established concepts in system safety. This review has appraised the methodology by consulting ML manufacturers to understand if it converges or diverges from their current safety assurance practices, whether there are gaps and limitations in its structure and if it is fit for purpose when applied to the healthcare domain. Through this work we offer the view that there is clear utility for AMLAS as a safety assurance methodology when applied to healthcare machine learning technologies, although development of healthcare specific supplementary guidance would benefit those implementing the methodology.

# 1. Introduction

The development of machine learning technologies for healthcare has seen significant growth in recent years. This can be witnessed through hundreds of ML-based medical devices gaining regulatory approval in the US and Europe between 2015 - 2020 [1]. However, these approvals are based on regulatory frameworks originally devised for traditional software, by which we mean, software that is constructed of rule-based algorithms for a specific task [2]. In comparison, ML software is not explicitly coded but instead developed by constructing a model that is learned through mathematical algorithms and training datasets to identify patterns which can be harnessed to make predictions on previously unseen data. This makes ML intrinsically data-driven and stochastic [3]. Therefore, possessing a fit for purpose safety assurance methodology for these emergent technologies will be a fundamental component in managing patient safety. We define safety assurance as "*all planned and systematic actions necessary to afford adequate confidence that a product, a service, an organisation or a functional system achieves acceptable or tolerable safety*" [4].

The novelty offered by ML has led to those with policy, safety and regulatory responsibility needing to appraise and update their existing regulatory frameworks and safety assurance routes through a series of projects and initiatives. Examples of such work include the UK Medicines and Healthcare Products Regulatory Agency (MHRA) changing its regulatory framework through the Software and AI as a Medical Device Change Programme [5] [6]; in England the Care Quality Commission (CQC), as part of their regulatory sandbox project on the use of ML as part of a service, highlighted in their findings a need to improve their methods to better regulate services which include ML [7] and; the US Food and Drug Administration (FDA) released a discussion paper in 2019 [8] from which they now have an action plan [9] to begin addressing their challenges. Similarly, the international standards community who develop standards that are often harmonised to regulation are embarking on similar projects to develop AI specific standards [10] [11]. While these nascent endeavours are being worked through a gap exists in how ML safety should be assured. The Assurance of Machine Learning for use in Autonomous Systems (AMLAS) [12] may offer a solution towards bridging this gap.

This paper presents an appraisal of the AMLAS methodology for its suitability as a safety assurance methodology to be utilised by digital health technology (DHT) manufacturers. Specific objectives were three-fold, (1) appraise how AMLAS converges or diverges from DHT manufacturers' existing safety assurance practices for ML; (2) to identify any gaps and limitations; and, (3) to identify key themes towards healthcare specific supplementary guidance for AMLAS. It should be noted, any divergence, gaps and limitations are scoped to the AMLAS and does not imply manufacturers are not compliant with their existing regulatory obligations.

The rest of the paper is organised as follows. Section 2 describes the methods. Section 3 describes in summary form the AMLAS methodology. Section 4 presents the results of how each AMLAS stage converges or diverges from existing manufacturer safety assurance practices and includes any gaps and limitations. Section 5 presents a discussion according to key themes and Section 6 concludes the paper.

## 2. Methods

### 2.1 Manufacturer Recruitment

Due to the nature of the expertise required to conduct this review, only those manufacturers deploying or ready to deploy their ML technologies were approached to participate. The researchers selected two manufacturers who fulfilled this criterion and are referenced throughout as **Manufacturer A** and **Manufacturer B**.

The core technology of each manufacturer is summarised below:

**Manufacturer A** – Developed a deep learning implementation of ML to assist radiologists with the decision to recall a patient if cancer is suspected on a mammogram. The current use case focuses on the ML being used as a second reader as currently all mammograms are read by two readers (i.e., a double reading workflow). It is envisaged the ML component will be integrated into existing breast screening pathways. This will involve them having a need to deploy their ML at various healthcare organisations' screening sites.

**Manufacturer B** – Developed a deep learning implementation of ML to identify patients who are at greatest risk of requiring non-elective care, with the ultimate goal of preventing these non-elective events from occurring. Once identified, patients are coached by the manufacturer's employees who work directly with patients, building up their health literacy and empowering them to take an active role in managing their health. This manufacturer both develops and deploys their technology and therefore is more analogous to a service provider.

### 2.2 Review Instrument

Prior to conducting the review, which was achieved through a series of workshops the lead researcher formulated a framework of questions based on each stage (1-6) of the AMLAS which needed DHT manufacturer input (see Appendix A for the entire framework). This approach ensured discussions focused on those salient points which needed exploration.

Table 1 illustrates a sample of the framework which showcases how the questions were presented and corresponding answers captured, all labelled with unique identifiers (e.g., MVA-1, MVA-1.Q1 etc). Each heading's purpose was as follows:

- **Key Discussion Point** – A key discussion point was extrapolated from the AMLAS directly quoting the text, where possible.

- **Review Question** – A question is presented linked directly to the key discussion point.

- **Review Answer - A** – Manufacturer A response.

- **Review Answer - B** – Manufacturer B response.

| Key Discussion Point | Review Question | Review Answer - A | Review Answer - B |
|---|---|---|---|
| **MVA-1:**<br><br>Model verification may consist of two sub-activities: test-based verification and formal verification. For every ML safety requirement at least one verification activity shall be undertaken. | **MVA-1.Q1:**<br><br>Do you agree with the key discussion point MVA-1 as a sensible approach to verification? | **MVA-1.Q1_A:**<br><br>At each new site we engage with we run a pass of our current ML model. Depending on performance we will calibrate with this new site data if needed. Additionally we have run and intend to run formal clinical investigations on performance in both a double reading and standalone workflow | **MVA-1.Q1_B:**<br><br>Yes, we currently verify our models with test-based verification |

**Table 1: AMLAS Review Questions Framework**

### 2.3 Data Collection

Online workshops consisting of 3 x 1hr were conducted with each manufacturer separately to collect the data. Each workshop was attended by the lead researcher, research team members and representatives from the manufacturer. Online document templates, as described in 2.2, were set up for each manufacturer. These were populated during the workshops and offline as some questions required input from the wider organisation.

### 2.4 Data Analysis

On completion of the data collection phase, the lead researcher combined the data into aggregate form to begin analysis. Since the sample size was small, specific coding was not utilised. Instead, each AMLAS stage has a convergence, divergence and gaps & limitation heading where each manufacturer's responses have been reported thematically. A convention adopted for clarity was as follows. If the combined responses reached agreement, they have been reported in the results section (4) as "both manufacturers …". Where there were no clear themes or agreement amongst the manufacturers, we reported them specifically as "manufacturer A/B …"

# 3. The Assurance of Machine Learning for use in Autonomous Systems

## 3.1 AMLAS Stages

AMLAS is a safety assurance methodology for autonomous systems which aims to integrate safety assurance during the development of ML components. For this reason, it is primarily constructed of iterative stages that resemble a typical ML engineering life cycle. There are six stages in total; (1) ML safety assurance scoping, (2) ML requirements, (3) Data management, (4) Model learning, (5) Model verification and (6) Model deployment. Figure 1 illustrates the six stages. There is a prerequisite to stage 1, which is to establish the system safety requirements that are used as an input into AMLAS. This is due to AMLAS taking a whole system safety approach even though the primary focus of the assurance methodology is on the ML component. AMLAS considers safety assurance to be meaningful if scoped as part of the wider system and operational context. The final box in Figure 1 labelled "Safety Case[1] for ML component" depicts the final artefact produced as part of implementing the entire methodology and should not be considered as a stage.

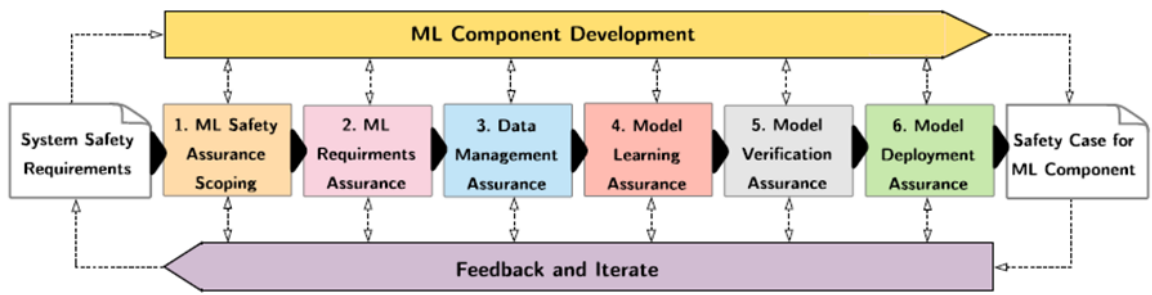

**Figure 1: Overview of AMLAS**

## 3.2 Processes & Argument Patterns

Each stage of AMLAS comprises of a process which takes in inputs that are used by the activities to produce stage outputs. Figure 2 is an example of the stage 1 process. Here, inputs A, B, C, D & F are all supplied to the activities 1 & 2, which output E & G.

---

[1] A safety case can be defined as "*a documented body of evidence that provides a convincing and valid argument that a system is adequately safe for a given application in a given environment*" [13]. The concept of a ML safety case in AMLAS harmonises with current healthcare safety standards (DCB0129 [14] & DCB0160 [15]) which include a requirement to produce a clinical safety case.

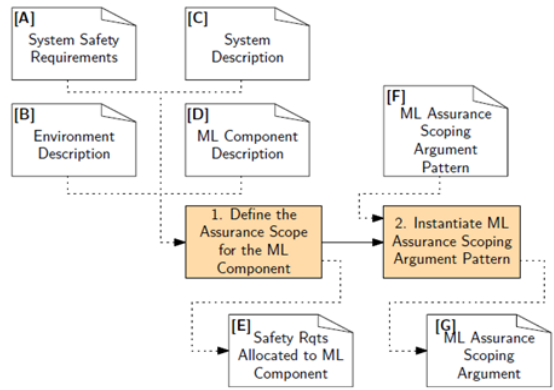

**Figure 2: ML Assurance Scoping Process (Stage 1)**

Furthermore, each stage as part of its process has an activity (e.g., see Figure 2, Activity 2) for instantiating a safety argument pattern[2] which references the artefacts of each stage. This gives implementers of the methodology clear guidance on how to present a safety argument informed by the safety work carried out at each stage. Figure 3 is an example of the safety argument pattern for stage 1. All six stages' arguments contribute to the final ML safety case.

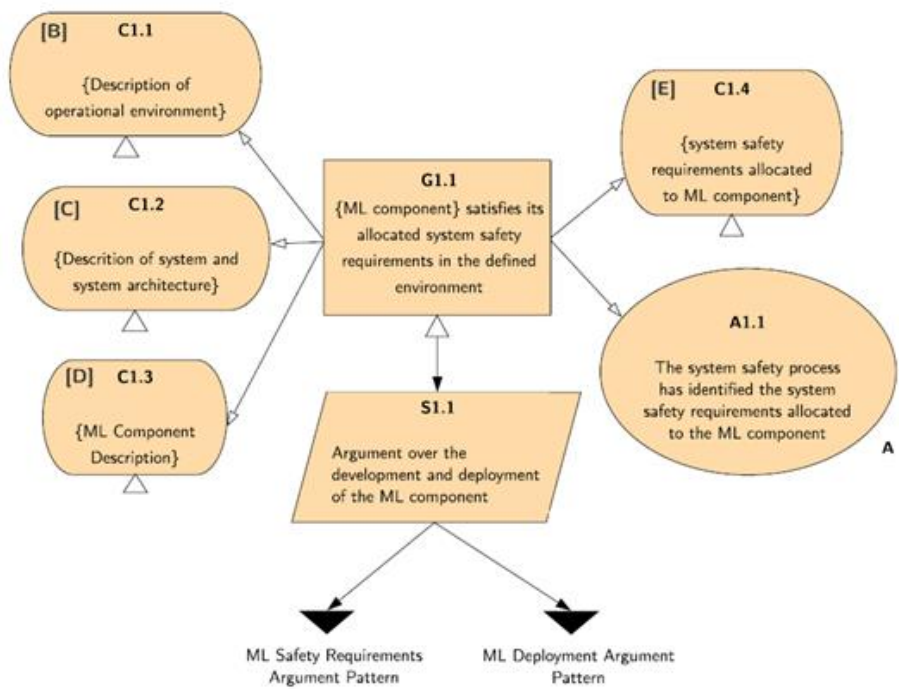

**Figure 3: Argument Pattern for ML Safety Assurance Scoping**

---

[2] A safety argument pattern explicitly illustrates the relationships between a safety claim, the context and the evidence required to satisfy the claim. AMLAS employs The Goal Structuring Notation Standard to formulate these patterns [16].

## 4. Results

This section presents the results of the review structured according to each stage of AMLAS.

### 4.1 Stage 1: ML Safety Assurance Scoping

Discussions at this stage centred around understanding current manufacturer practice of safety assurance using a whole system safety approach and the feasibility of addressing the inputs (A, B, C, D & F), activities (1 & 2) and producing outputs (E & G) as prescribed by AMLAS.

**Convergence**

Both manufacturers understood and accepted the concept of carrying out a system safety assessment prior to the ML safety assurance scope.

Manufacturer A explained how the inputs of this stage are already in part addressed through harmonised standards, clinical workflow plans, risk assessments and deployment methodologies. In addition, manufacturer A stated,

> *"In depth understanding of the breast screening system was essential when building the integrations needed for the ML component to fit seamlessly into the current clinical workflow"*

Manufacturer B, being both a manufacturer and deployer of the ML component understood the need to scope the safety of the ML from a system-level. They expressed how they had already scoped the safety of the ML according to decisions it made and how that impacted the wider clinical pathway.

Both manufacturers were confident they would be able to address the inputs and activities leading to the desired outputs as prescribed by AMLAS at this stage.

**Divergence**

Both manufacturers stated their ML component safety scope was gathered from its binary output and how that decision affected the wider system. This invariably led conversations towards performance metrics. This involved assessing the safety of the ML mostly against performance metrics, which if acceptable to the manufacturer, would translate into the ML's contribution to the wider system as being safe. However, if performance metrics are not context specific enough (e.g., different patient types may require specific metrics), they could contribute towards compromising safety of the wider system. Further guidance for manufacturers of how to explicitly consider the ML safety assurance scope linked to a wider system safety assessment would be beneficial.

In practice, both manufacturers were producing content similar to the inputs A, B, C, D for their regulatory tasks and internal quality assurance routes, although not always as distinct artefacts as defined per AMLAS at this stage.

**Gaps and Limitations**

Discussions highlighted the importance of including qualified healthcare professionals (HCPs) as subject matter experts due to the nature of how integral they are in healthcare pathways. AMLAS allows for this and provides some guidance through the notes and examples. However, explicit identification of where they should be involved would provide benefit.

**4.2 Stage 2: ML Requirements Assurance**

Discussions at this stage centred around understanding current manufacturer practice of assigning safety requirements to the ML component.

**Convergence**

It was clear from discussions, both manufacturers understood and made use of performance and robustness metrics for safe operation of the ML. However, their assignment was with an implied system level thinking, e.g., ML false negative equals potential towards patient harm.

Manufacturer A performance metrics included "*recall rate (RR), cancer detection rate (CDR), sensitivity, specificity assessed as part of each deployment to ensure performance on a per site basis*". Robustness of the component was addressed through "*any cases that the ML cannot read i.e., technical recalls, not sufficient images etc, are not processed through the tool*".

Manufacturer B made use of "*no false negatives (FN) associated with extreme events such as mortality*" and an "*area under the curve (AUC) of >0.80%*". Furthermore, measuring the ML performs equally across differing patient attributes was a key performance concern for this manufacturer. Regarding robustness, they augmented missing data by accessing alternative data sources which in turn allowed the model to continue with desired outputs.

**Divergence**

This paragraph repeats the corresponding comment in the above stage. Both manufacturers were not explicitly cascading the system level safety requirements to a ML safety assurance scope and then assigning ML safety requirements, as prescribed per AMLAS.

**Gaps and Limitations**

Interpretability was the only area which was discussed as a potential addition to the types of ML safety requirements. AMLAS allows for additional ML safety requirements to be added. Manufacturer B raised an interesting point of "*… maybe to distinguish between interpretability vs explainability*". A way forward would be to consider them under the broader term "transparency", which expands further to include attributes such as fairness etc. This will need to be explored further by the research team.

### 4.3 Stage 3: Data Management Assurance

Discussions at this stage centred around understanding the concept of data requirements which are sufficient to allow for the ML safety requirements to be encoded as features against which the data sets to be produced in this stage may be assessed.

**Convergence**

Both manufacturers made use of well-established data spitting methods. Specifically, the data set is split into a development (training) set and a verification/validation set.

Manufacturer A stated:

> "*data set for our large scale Trial 2 was split into 2 sets; trial set vs development set. Where the development set was used for machine learning and the trial set was used to verify/evaluate the tool against. Reason for this was to evaluate on before unseen data*"

Manufacturer B stated:

> "*The model is trained on 75% of the total data, where 25% of the data is held-out for validation and reporting purposes.*"

Terminology differs amongst manufacturers in a similar fashion to the ML community, although the concepts hold.

**Divergence**

Currently, both manufacturers obtain their data sets from real-world settings. Primarily, this data is obtained from partnered healthcare organisations. This does provide data of the intended target population but comes with associated biases and limitations. For example, real-world data does not necessarily ensure the entire population is present in the data set as this is dependent on attendance or having access to data of all types of patients. However, it is not always necessary for all patient types to attend or be present in the data set, in order for a model to generalise due to the prevailing concept of ML (i.e., to learn patterns from training data which can be applied to unseen data). However, this can only be assessed from rigorous evaluation.

The challenge here is for manufacturers to develop explicit safety requirements based on the AMLAS data attributes, "*relevant*", "*complete*", "*accurate*" and "*balance*", to assess the safety of the ML component when generalising for previously unseen population.

**Gaps and Limitations**

This point formulated below as a question was explicitly raised by the lead researcher and will continue to be explored further as no clear consensus was reached.

Will the AMLAS data attributes suffice towards data safety requirements associated with healthcare data? For example, under which attribute would data distribution drift/shift be included? It should be noted, AMLAS does allow for additional attributes to be added.

### 4.4 Stage 4: Model Learning Assurance

This stage focuses on developing the machine learnt model using the development data obtained in the previous stage such that the allocated ML safety requirements are satisfied. However, since this was not explicitly occurring as part of current assurance practices the discussions involved understanding how it could be achieved and managed by the manufacturer.

**Convergence**

Regarding the activities as prescribed by AMLAS, both manufacturers create models and test them based on performance metrics and robustness requirements in an iterative manner. This feedback loop informs optimisation techniques which leads to the candidate model for deployment. This is similar in nature to the requirements of AMLAS, except they would need to align existing model learning to ensure ML safety requirements are met based on the data set.

The practice of discussing trade-offs is followed by both manufacturers as it is a fundamental aspect of acquiring the best fit model. Manufacturer A pointed towards aspects of their regulatory requirements as a factor that drove their discussions but were limited in what they could state due to Intellectual Property (IP).

Manufacturer B mentioned having a need to maintain an area under the curve (AUC) >0.80%. This informed their decision of how regular the data acquisition cycle should run to maintain their metrics leading to trade-offs between what can be practically achieved with partnered organisations and internal metrics.

**Divergence**

Both manufacturers are expecting specific metrics to be met in order to provide evidence of the viability of their ML component. Hence, it is implied safety of the component will be justified from the metrics. However, these metrics will need to be explicitly linked to the wider system safety requirements to be meaningful. Manufacturer A exemplifies this thinking by stating:

> *"As discussed, in the context of our product it is difficult to separate improved performance with improved safety; i.e., if the model is more accurate with its binary decision it is safer. However, when looking at safety as a whole it comes down to more than just ML performance metrics when implemented into a wider system and this is covered through robust deployment methodologies and novel workflows using AI".*

Manufacturer B concerns were:

> "*ML engineers want to deploy models and safety could be considered as additional work that takes them away from their core role so I believe the form of communication should be in the easiest/least time-consuming way to follow.*"

Both points of view highlight that ML engineering teams are immersed in solving complex problems that are intrinsically linked to metrics. The research team will need to address explicit inclusion in any guidance of the potential pitfalls of linking metrics to safety. To elaborate, more thought will need to be given on how specific metrics translate towards safety for a specific use case and whether the metrics are applicable to all types of patients and rare conditions.

The Model Development Log has a pivotal role as per the AMLAS. This was not something that was currently used as part of a safety process. However, manufacturer B provided a sensible approach to how it could be utilised with the time constraints already placed on ML engineering teams:

> "*The project/workstream owner to map out the process, where the ML engineers have to mark which stages will have the highest impact on the performance of the model and all changes made to these key pivots are to be logged*"

As mentioned, "*… impact on the performance of the model and all changes made to these key pivots are to be logged*" could be explicitly linked to how it affects the safety requirements. Any associated work that needs to be conducted could be recorded in the model development log.

**Gaps and Limitations**

A specific inclusion criterion in the argument could be to provide explicit justification as to why specific metrics are chosen for optimisation of performance particularly to enhance the safety. This is best expressed by manufacturer B as:

> "*It won't be superfluous to also describe what it will mean if one of the metrics is chosen for optimisation. E.g., if you choose false negatives that means you will have higher sensitivity and lower positive predictive value. Again, helping to build a common understanding on related but often assumed similar metrics*"

**4.5 Stage 5: Model Verification Assurance**

Discussions at this stage were centred on verification of the ML model based on current manufacturer practice in comparison to specific AMLAS verification requirements.

**Convergence**

Of the two available options in AMLAS to verify a model, both manufacturers made use of test-based verification. Manufacturer A conducted their verification by stating:

> "*At each new site we engage with we run a pass of our current ML model. Depending on performance we will calibrate with this new site data if needed. Additionally, we have run and intend to run formal clinical investigations on performance in both a double reading and standalone workflow*"

Based on AMLAS guidance, the verification data not being made available to the development team was seen as a sound concept by the manufacturers, however manufacturer B did state,

"*this is tricky as additional data for verification might be difficult to obtain*". The lack of good quality data available in ample quantities is an ongoing challenge for the machine learning community. A further practical challenge involves shielding the verification data from the development team.

**Divergence**

No divergence to report.

**Gaps and Limitations**

One particular area that was discussed and could be included is an explicit requirement to verify against bias/fairness of the model for sub-groups of patients. These sub-groups would need to be proactively identified as those that are most at risk of being unfairly discriminated against, depending on the use case being executed.

Manufacturer B explains this by stating:

> "*… additionally models could also be tested/verified on their bias/fairness for particular unprivileged groups. This is particularly relevant in healthcare ….*"

**4.6 Stage 6: Model Deployment Assurance**

AMLAS is written from a development (ML engineering) perspective, therefore it is fair to state that within its guidance much of the responsibility of deploying safely is placed on the manufacturer. However, if the technology is to be deployed into a third-party environment, responsibility of deploying safely becomes more a joint responsibility between the manufacturer and healthcare organisation. This aspect was given emphasis through discussions at this stage.

**Convergence**

Deployment was seen as a multi-stage activity involving integration into the existing hardware & software infrastructure and the corresponding clinical pathway to test safety requirements satisfied during pre-deployment stages continue to be satisfied. In line with this, manufacturer A is taking the approach of deploying a model which is frozen and then monitored. Any subsequent changes that are required that affect the wider system safety requirements are discussed and approved by a multi-disciplinary team. Routine monitoring and modification of the model will be necessary to continue meeting safety requirements. Regarding this matter, Manufacturer B stated, we should be asking prior to any modifications or updates "*why are we changing the model?*". Furthermore, keeping "*track of what changed and its impact on output*" should be logged which aligns with the AMLAS.

With regard to specific safety assurance process and documentation, Manufacturer A are complying both with the CE marking process (Medical Device Regulation) [17] and the Health Information Technology (HIT) standard for manufacturers, DCB0129 [14]. The evidence is being developed as a technical file and safety case, respectively. In addition, Manufacturer A has instructed the healthcare organisations to complete the DCB0160 (HIT standard for the deployment and use of HIT systems) [15]. Involving the healthcare organisation and their clinicians in this process is seen as vital as manufacturer A stated, "*our phased deployment has vast involvement from the deploying organisation, particularly the clinicians*".

**Divergence**

AMLAS places significant responsibility on those who follow the methodology to complete logs (development, error, etc). This is widely used in engineering, however it was felt from a practical perspective, completing logs should in the most be automated rather than human led. This is not a clear divergence from AMLAS, more a clarification of opinion.

**Gaps and Limitations**

Deployment in itself will provide safety assurance linked to temporality, whereas routine monitoring and modification of the ML will be needed to continue satisfaction of safety requirements. Therefore, we see a need for routine monitoring and modification to be expanded into additional stages as they are not covered in sufficient detail at stage 6. Within the healthcare domain, on-going monitoring and modification are crucial stages where safety assessments continue as part of regulatory compliance and HIT standards.

## 5. Discussion

Five conceptual themes emerged from this deep review of the AMLAS which are discussed below as, (1) ML safety as part of whole system safety; (2) explicit inclusion of HCPs leading to richer safety assessments; (3) mapping the contribution of performance metrics & soft constraints towards the ML safety profile; (4) data management processes to satisfy safety requirements; and, (5) apportioning roles and responsibilities between the manufacturer and deploying organisation to maintain safety requirements of the ML in live operation.

The manufacturers, depending on the nature of how they build and deploy ML, are considering the safety of their product as part of the wider system. Based on the classification of their technology, this involved formal safety assessments through their work associated with regulatory compliance routes or internal quality & safety approaches. Manufacturers understood the concept of safety assurance from a whole system approach which is decomposed to specific ML safety requirements. This concept is in part addressed through their existing processes although not from the initiating phases of development as prescribed by AMLAS. However, both manufacturers agreed they could comply with ML safety from a whole system approach. The proposed future work (healthcare specific supplementary guidance for AMLAS) should include current methods in place that allow for derivation of system safety requirements.

AMLAS throughout its guidance makes reference to the inclusion of experts in the safety assurance process. This was a recurring discussion point in the workshops of where qualified HCPs should be included to provide much needed clinical expertise in safety assessments. The benefits of their participation is self-evident due to being subject matter experts bringing numerous benefits, one of which is to contribute to those areas which need human factors to be considered as part of safety assessments, such as automation bias, handover, etc. A key finding from this work is to consider where HCPs shall/should be included explicitly in the AMLAS assurance process and argument patterns. It should be noted, the group recognised best-practice to have HCPs included throughout, although the reality of obtaining their time can be challenging, therefore leading towards selective inclusion or creation of dedicated roles.

Metrics are fundamental to how the manufacturers assess the safety of their ML. Discussions identified common metrics used were sensitivity and specificity, expressed as Area Under the curve of the Receiver Operating Characteristic (AUROC), and benchmarked recall rates to satisfy internal target performance criteria which linked to implied safety. However, having high-performance levels is only part of the solution to assuring the safety of the ML. "*Soft constraints*", such as transparency (e.g., interpretability, explainability, etc), will need to be considered from the viewpoint of how they impact human factors [18] and should potentially be explicit criteria as part of safety assessments. One specific standard, but by no means the only one, ISO 62366: Application of Usability Engineering to Medical Devices can assist the thinking required to address soft constraints. Furthermore, manufacturers will need to consider what trade-offs would need to be made to the performance of the ML to achieve these soft constraints.

As expected, manufacturers made use of well-established data management techniques and were splitting their data sets as per current methods. However, they were not setting safety requirements of their data to be relevant, complete, accurate & balanced as per AMLAS guidance. This does not imply manufacturers are not making use of their own techniques and the concepts were accepted as being extremely important and integral to any ML project. A question that arose from this which requires further research was, under which attribute/s would data distribution drift/shift be included?

Currently, both manufacturers obtain their data sets from real-world settings (partnered healthcare organisations) which come with associated biases and limitations. Therefore, to comply with AMLAS the challenge for manufacturers here is more to change their mindset in how they currently approach data management and the assignment of safety requirements to data which produce models that satisfy ML safety requirements. This is by no means a simple task and the hope is future guidance will help towards this goal.

Finally, any argument for having an appropriate data set should present why it is sufficient to produce models that generalise for previously unseen populations (i.e., patient subgroups). This is particularly relevant to healthcare as models will eventually be deployed on thousands of patients coming from diverse and somewhat fluid populations.

During the engineering of a ML component heavy emphasis is placed on its performance accuracy and for that to hold when deployed in a real-world live setting. This can often translate as a safe product which is not the case as there are other factors, such as transparency, which will be just as integral in safe deployment. This responsibility, as per AMLAS, is apportioned to the manufacturer. However, if deploying at a third-party site, the deploying organisation should be fully involved with the manufacturer in safely integrating the ML into their existing hardware/software infrastructure and clinical pathway. AMLAS currently does not explicitly include the deploying organisation to be involved in the deployment phase, nevertheless it is flexible enough for its inclusion. Furthermore, routine monitoring is crucial to satisfying safety requirements as this stage allows for data to be gathered on safety requirements being maintained and justification for when change is required. AMLAS does not include any stages beyond deployment and therefore, additional stages or safety argument updates will need to be considered by the research team as per their future work.

### 5.1. Limitations

Every effort was made to recruit and work with DHT manufacturers with ML technologies which are deployed or in the process. While this was achieved, having two manufacturers does limit the review to specific technologies, scenarios and working practices. A greater number of manufacturers may have yielded further conceptual themes of interest.

## 6. Conclusion

Assuring the safety of ML-based technologies has never been more pressing with the current upward trend of ML technologies gaining regulatory approval through frameworks originally devised for traditional software. As those organisations with policy, regulatory and safety responsibility continue to reform their frameworks, methodologies such as the AMLAS need to be appraised and evaluated to assess whether they are fit for purpose as safety assurance methodologies alongside regulation. This work has concluded the methodology to be one that is appropriate to be applied in the healthcare domain with additional healthcare supplementary guidance.

**Acknowledgements**

This work is part of work package 2 of the Safety Assurance FRamework for Machine Learning in the Healthcare Domain (SAFR) project, funded through the Assuring Autonomy International Programme (AAIP), which is overseen by the University of York and Lloyd's Register Foundation.

# Appendix A - AMLAS Review Questions Framework

## ML Safety Assurance Scoping (MLSAS)

| Key Discussion Point | Review Question | Review Answer - A | Review Answer - B |
|---|---|---|---|
| **MLSAS-1**: This stage defines an ML component as: **An ML component comprises an ML model, e.g. a neural network**, that is deployed onto the intended computing platform | **MLSAS-1.Q1**: Do you agree with the definition of a ML component? Please choose from **Yes/ No/ Unsure** and provide a justification for your chosen option. | **MLSAS-1.Q1_A**: - confirm agree. The ML component of the product is the neural network that outputs a binary decision of recall or no recall. This is integrated into a standard clinical workflow | **MLSAS-1.Q1_B**: Yes |
|  | **MLSAS-1.Q2**: Is your product a combination of traditional IT and ML? Provide a brief description | **MLSAS-1.Q2_A** - The product is an ML component that runs in the cloud with integrations into local breast screening IT systems which includes the ability to query PACS systems and intergate and write into NBSS | **MLSAS-1.Q2_B**: Whats the definition of traditional IT: <br> - series of processes that comes before ML <br> - actions based on set of pre defined conditions (which we're not) |
| **MLSAS-2**: The safety requirements allocated to the ML component shall be defined to control the risk of the identified contributions of the ML component to system hazards. This shall take account of the defined system architecture and the operating environment. At this stage the requirement is independent of any ML technology or metric but instead reflects the need for the component to perform safely | **MLSAS-2.Q1**: During the early stages of development, would you have been able to obtain information and describe the system and its architecture in sufficient detail? Please choose from **Yes/ No/ Unsure** and provide a brief explanation of how and who would do this | **MLSAS-2.Q1_A**: Yes, software requirements prior to coding. Medical device processes - IEC62304. Medical device software lifecycle process, frame how you do medical software engineering and quality control. This process would be a combination of ML, engineering teams and QARA (Quality Assurance and Regulatory Affairs) | **MLSAS-2.Q1_B**: Daily feed from hospital -> ML -> risk scores -> patient screening by clinical coach <br><br> The datasets that the prediction ML model needs for daily scoring is dependent on the system and its architecture. |

| | Question | Answer A | Answer B |
|---|---|---|---|
| | **MLSAS-2.Q2:** During the early stages of development, would you have been able to obtain information of the operating environment that your technology was deploying into? Please choose from **Yes/ No/ Unsure** and provide a brief explanation of how and who would do this | **MLSAS-2.Q2_A:** the ML is run in the cloud but integrates the PACS and the breast screening systems local to the site of deployment. Indepth understanding of the breast screening system was essential when building the integrations needed for the ML component to fit seemlessly into the current clinical workflow. MDDS - medical device data system acts as a linkage between the medical device with the ML component and the PACS/NBSS. The ML component runs in the cloud and interacts with the systems via MDDS | **MLSAS-2.Q2_B:** Before developing the ML model, the team would obtain information about:<br>- data refresh (key in deployment since we need to identify at-risk patients as soon as possible, thus we require data to be refreshed as soon as possible/is available),<br>- pseudo anonimisation,<br>- data period (the model is trained on at least 3 years worth of hospital data)<br>- availability of the team that manages the environment in which the model will be deployed |
| | **MLSAS-2.Q3:** Are you be able to describe the ML component in sufficient detail, including explicitly the intended use? Please choose from **Yes/ No/ Unsure** and provide a brief description. | **MLSAS-2.Q3_A:** Yes. Included in CE marking technical file including the intended use | **MLSAS-2.Q3_B:** A generalised linear model (GLMnet) that uses routinely collected data to predict risk of care consumption, defined by the patient's predicted likelihood of staying in the hospital for 3 bed days or more. |
| | **MLSAS-2.Q4:** During the early stages of development would you have been able to explain the role of the ML component in the system? Please choose from **Yes/ No/ Unsure** and provide a brief explanation of the role of the ML | **MLSAS-2.Q4_A:** Yes - ML component is the key part of the intended use of the medical device, captured in product requirement specifications and software requirement specifications. Informed by IEC. Breaking the product down and then the software coding requirements to each of these parts | **MLSAS-2.Q4_B:** Yes, the ML takes routinely collected data to identify high cost high need patients. This is done by estimating the likelihood of a patient spending 3 unplanned bed days or more in hospital. |
| | **MLSAS-2.Q5:** During the early stages of development would you have been able to conduct a system safety assessment (hazard identification & risk analysis)? Please choose from **Yes/ No/ Unsure** and explain briefly how and who would do this. | **MLSAS-2.Q5_A:** Yes. Undertaken as part of the CE marking. QARA team - collaboration effort | **MLSAS-2.Q5_B:** Yes, by looking at the model's performance metrics. In particular, we use false negative rate to assess if any false negatives are associated with extreme events such as mortality, or the misclassification is acceptable. |
| | **MLSAS-2.Q6:** Would you be able to explicitly consider the human (human factors) as part of the system safety assessment (handover, automation bias)? Please choose from **Yes/ No/ Unsure** and provide some rationale for your answer | **MLSAS-2.Q6_A:** Yes. Undertaken as part of the CE marking | **MLSAS-2.Q6_B:** Yes - but it is unclear where to draw the line and if there is no human factor what should be in place.<br><br>For example if we focus on false positives & false negatives which can be easily covered by clinical staff reviews, should there be mechanisms to track this? |
| | **MLSAS-2.Q7:** Would you be able to allocate safety requirements to the ML component from the system safety assessment? Please choose from **Yes/ No/ Unsure** and provide a brief explanation of your approach | **MLSAS-2.Q7_A:** Yes, the ML component is what supplies the binary recall/no-recall decision for a breast screening participant, therefore any safety considerations related to this decision are attributed to the ML component. Also part of risk review, evaluated and risk controls and mitigations are implemented | **MLSAS-2.Q7_B:** Yes. Our safety requirements fall out from our focus on patients with spiralling levels of unplanned care, therefore performance metrics are key (AUC and FN and FP). Additionally for the health domain, we recognise two more important factors - helping patients to live longer & ensuring outcomes dont differ by deprivation/ethnicity |

| | MLSAS-2.Q8: Would you be able to setup a system level risk acceptance criteria? Please choose from **Yes/ No/ Unsure** and briefly describe the criteria | MLSAS-2.Q8_A: Yes - part of the risk analysis process (Greenlight Guru - QMS). V&V (verification and validation), tests of the software are undertaken in terms of full system test/workflow | MLSAS-2.Q8_B: Unsure - our current checks are based on Yes / No rather than specific thresholds and acceptance criteria |
|---|---|---|---|

# ML Safety Requirements Assurance (MLRA)

| Key Discussion Point | Review Question | Review Answer - A | Review Answer - B |
|---|---|---|---|
| **MLRA-1:** Develop the machine learning safety requirements from the allocated system safety requirements. ML safety requirements shall be defined to control the risk of the identified contributions of the ML component to system hazards. | **MLRA-1.Q1:** What approach would you take to allocating ML safety requirements from the system safety requirements? | **MLRA-1.Q1_A:** The ML is responsbile for the binary recall/no recall decision - safety considerations associated with this decision are part of the ML component. Whereas the wider system is responsbile for the tool's integration into hospital systems and workflows, therefore any risks associated with this are less dependent on the ML component. Full system V&V is conducted as part of the ML quality assurance. Wider clinical workflow system safety is ensured as we do local level service evaluation and clinical investigations | **MLRA-1.Q1_B:** |
| **MLRA-2:** The ML safety requirements shall always include requirements for performance and robustness (see example 7) of the ML model in relation to system safety requirements. | **MLRA-2.Q1:** What performance metrics do you use at present to assure your ML component operates safely? | **MLRA-2.Q1_A:** recall rate (RR), cancer detection rate (CDR), sensitivity, specificity assessed as part of each deployment to ensure performance on a per site basis. Medical device V&V as part of quality assurance is done | **MLRA-2.Q1_B:** No false negatives (FN) associated with extreme events such as mortality<br>AUC > 0.80%<br>Ensure the ML performs equally between patients of different ethnicities/deprivation levels<br>Mortality and ethnicity are particularly relevant for health (i.e. no patients have passed away in the FNs and that differences by social demographics are captured and actioned on) |
| | **MLRA-2.Q2:** At present how do you ensure your ML component is robust and operates safely when it encounters a situation it can't deal with? | **MLRA-2.Q2_A:** any cases that the ML cannot read i.e. technical recalls, not sufficient images etc. are not processed through the tool. System operations are being built to ensure this is flagged or processed alternatively through the clinical workflow. Any dataset that doesnt ahdere to the ML's data learning spec is not analysed and event is recorded in error logs and failure logs for user interrogation. | **MLRA-2.Q2_B:** Comment: Our metrics for robustness is an AUC of >0.8. We don't have a threshold for FN or FP yet, but safe operation revolves around avoiding harm and monitoring whether patients with unplanned care events who haven't been identified by the prediction model have had extreme events.<br><br>The model performed fairly well during the COVID-19, but we might want to built explicit tests for any future models that will factor in significant changes in the environment.<br><br>We also ensure robustness by using alternative data sources like A&E (and not just Inpatient) which helps fill in the information gap.<br><br>If there is missing data, this results in lower risk scores. |
| **MLRA-3:** 'Soft constraints' such as interpretability may be crucial to the acceptance of an ML component especially where the system is part of a socio-technical solution. All such constraints defined as ML safety requirements must be clearly linked to safety outcomes. | **MLRA-3.Q1:** Should interprebility of the ML component be considered as an explicit safety criteria? Please choose from **Yes/ No/ Unsure** and provide some rationale for your answer | **MLRA-3.Q1_A:** - yes, should be considered from a usability engineering perspective and binary recall/no recall has been designed to minimise misinterpretation or misuse | **MLRA-3.Q1_B:** Comment: Yes, but we have to balance benefit vs cost. E.g. imaging is extremely ineffective with non-deep learning methods so having lower levels of interpretability should be ok. However, requirements for developers to show incremental benefits of different models could be made compulsory.<br><br>A good idea will be maybe to distinguish between interpretabiliy vs explainability |
| **MLRA-4:** Validate the ML safety requirements against the allocated safety requirements, the system and software architecture and operational environment. | **MLRA-4.Q1:** At this stage, how would the ML safety requirements be validated against the allocated system safety requirements? E.g. AMLAS states common approaches as **domain expert reviews** and **simulations** | **MLRA-4.Q1_A:** System tests are performed between the ML component and Medical Device Data System (MDDS) and full system testing is done on site by site level. Clinical workflow system impacts are measured through the deployment phases and formal clinical investigations. Technical workflow systems are tested in both test environments and also eventually within a prospective clinical investigation | **MLRA-4.Q1_B:** Comment: This is a very good point to make explicit and most importantly to think about whether the ML is deployed into an existing or new clinical pathway.<br><br>If it's an existing pathway a majority of the safety requirements will fall from the AI model. However, if it is a new pathway it could be a compulsory requirement to embed with safety requirements of the operational and deploying environment. |

# Data Management Assurance (DMA)

| Key Discussion Point | Review Question | Review Answer - A | Review Answer - B |
|---|---|---|---|
| DMA-1: Develop data requirements which are sufficient to allow for the ML safety requirements to be encoded as features against which the data sets to be produced in this stage may be assessed. | DMA-1.Q1: Gather ML developer opinion on DMA-1 | DMA-1.Q1_A: Data is collected from a broad set of partners to ensure there is sufficient opportunity to train on data from various hardware vendors and demographics | DMA-1.Q1_B:<br>- at least 3 years worth of data covering all A&E and Outpatient attendances and Inpatient admissions<br>- activity and medical diagnosis data is a minimum requirement<br>- patient characteristics should be included (age, gender, deprivation index) |
| DMA-2: ML data requirements shall include consideration of the relevance, completeness, accuracy and balance of the data | DMA-2.Q2: What other considerations need to be taken into account? | DMA-2.Q2_A: | DMA-2.Q2_B: Data missingness should be <5% especially for fields which are of highest importance or for models that are built on target variables where the target attribute is less than <10% (e.g. the thresholds we use marks ~92% of all inpatient spells as 0) |
| DMA-3: This shall include three separate datasets: Development data [N], Internal test data [O] and Verification data [P]2. The first two of these sets are for use in the development process (Stage 3) whilst verification set is used in model verification (Stage 4). | DMA-3.Q1: Currently how do you split data sets? | DMA-3.Q1_A: - data set for our large scale Trial 2 was split into 2 sets; trial set vs development set. Where the development set was used for machine learning and the trial set was used to verify/evaluate the tool against. Reason for this was to evaluate on before unseen data | DMA-3.Q1_B: The model is trained on 75% of the total data, where 25% of the data is held-out for validation and reporting purposes. |
| | DMA-3.Q2: How are they used for verification? | DMA-3.Q2_A: model verification is done by running the ML on the before unseen data set to assess performance against key relevant metrics i.e. sensitivity and specificity | DMA-3.Q2_B: Outputs validated by clinical staff at the point of training, and in deployment through fortnightly review sessions. The feedback loop between technology and front-line staff is the foundation of AICC. |
| DMA-4: The generation of ML data will typically consider three sub-process: collection, pre-processing and augmentation. | DMA-4.Q1: How are data sets currently generated? Augmented? | DMA-4.Q1_A: - data sets are obtained through extraction of retrospective data from clinical partners. This consists of an extraction of all historic digital mammograms (DICOM files) and their associated clinical information as ground truthing | DMA-4.Q1_B: Datasets are combined to form a holistic, cross-department view of the patient's medical and care utilisation history. Time-dependent features are created to represent patients' historical activity and conditions.<br><br>The source of data is the hospital, we pre-process and create the features.<br><br>Re augmentation - i think we can think about it from an internal view (what the model generates) and the direct feedback (From the clinical coach screening). And also external systems such as community recorded mortality (vs just hospital) or Navigator to further augment the impactability question |

| Requirement | Question | Answer A | Answer B |
|---|---|---|---|
| **DMA-5:** Validation of data relevance shall consider the gap between the samples obtained and the real world environment in which the system is to be deployed | **DMA-5.Q1:** How is data relevance currently managed? | **DMA-5.Q1_A:** - data obtained is from real world settings and is therefore completly relevant to the environment the tool will be deployed in | **DMA-5.Q1_B:** Our models are always trained with data from the population where we are deploying our system in. This ensures the model is tuned to the particular consumption patterns and chronic conditions of the population. |
| **DMA-6:** Validation of data completeness shall demonstrate that the collected data covers all the dimensions of variation stated in the ML safety requirements sufficiently. | **DMA-6.Q1:** How do you currently assure data completeness? | **DMA-6.Q1_A:** - data sets obtained merge together DICOM files and associated clinical information. The fields of associated clinical information required are defined through our de-identifcation process and those required for learning/analysis are kept just justifications | **DMA-6.Q1_B:** Drop rows where crucial information is missing (patient ID, age, record ID)<br><br>**Comment:** I think we definitely check that the dropped rows are not >5% of the total data? |
| **DMA-7:** Validation of data balance shall consider the distribution of samples in the data set | **DMA-7.Q1:** How is the data balanced? Equal number of data sample? | **DMA-7.Q1_A:** - data is obtained from real world data sets and is therefore representative of real world distributions, with any real world limitations existing | **DMA-7.Q1_B:** Class imbalance is handled by applying weights that impose a heavier cost when errors are made in the minority class |
| **DMA-8:** Validation of data accuracy shall consider the extent to which the data samples, and meta data added to the set during pre-processing (e.g. labels), are representation of the ground truth associated with samples. | **DMA-8.Q1:** How do you assure data accuracy? | **DMA-8.Q1_A:** - data set accuracy are limited by the accuracy of local medical records. Ground truth labels are obtained from historic entries into systems e.g. pathology results and historic human read of the images and are merged with the image files to create a holisitic case view | **DMA-8.Q1_B: Comment:** We don't do that at the moment partially due to data access, but national stats provided by NHS Digital could be a way to externally verify this. At the moment we can only verify A&E attendances and NELs for a deploying hospital but that is not sufficient. |
| **DMA-9:** Verification data is gathered with the aim of testing the models to breaking point. | **DMA-9.Q1:** What factors would need to be taken into consideration to gather a verification dataset that tests models to breaking point? | **DMA-9.Q1_A:** | **DMA-9.Q1_B: Comment:** I think the key will be to capture what data was used at the point of training and then test against real data that doesn't have the same distribution (e.g. a model trained on hospital data before 2020 vs COVID-19) |

# Model Learning Assurance (MLA)

| Key Discussion Point | Review Question | Review Answer - A | Review Answer - B |
|---|---|---|---|
| **MLA-1:** The creation of an ML model starts with a decision as to the form of model that is most appropriate for the problem at hand and shall be most effective at satisfying the ML safety requirements. This decision may be based on expert knowledge and previous experience of best practice. | **MLA-1.Q1:** How should it be communicated to ML engineers that they need to create a model which solves the problem and **satisfies safety requirements?** | **MLA-1.Q1_A:** Struggling with this question - as satisfying the safety requirements i.e. having sufficient sens/spec/CDR/RR is also the goal of actually buidling a model that works - I don't necesseciarly see them as two separate things? As discussed, in the context of our product it is difficult to separate improved performance with improved safety; i.e. if the model is more accuarte with its binary decision it is safer. The larger safety concerns come into play when it is in a clinical workflow and there are other factors in play; here we are thinking more about the robustness of the model and the integrations into a workflow. Examples of this are cases that the ML cannot read e.g. technical recalls of more than 4 images, or women with implants - a process needs to be in place to ensure these cases take a route not through our product. To note that these are also exclusion criteria on our CE marking authorised uses. | **MLA-1.Q1_B:** Comment: ML engineers want to deploy models and safety could be considered as additional work that takes them away from their core role so I belive the form of communication should be in the easiest/least time consuming way to follow. **However**, it might be down to the project owner to define the overall requirements and then translate to the technical team.<br><br>Having done some evaluations (or benefits tracking) and testing will definitely help supplement the gathering of requirements. |
| **MLA-2:** In creating an acceptable model it is important to note that it is not only the performance of the model that matters. It is important to consider trade-offs between different properties such as trade-offs between cost of hardware and performance, performance and robustness or sensitivity and specificity. | **MLA-2.Q1:** How are tradeoffs discussed and agreed amongst the development team? | **MLA-2.Q1_A:** Operating points established as per CE marking (will dig into this with ML team on what this decision was based on) | **MLA-2.Q1_B:** Trade-offs discussed include:<br>data period required for training;<br>AUC should achieve >0.8;<br>maximum data latency that ensures performance does not decrease;<br><br>Comment: We didn't do it necessarily from a safety aspect, but we calculated the time between flagged as a risk & crossing the threshold to inform the question about data latency? |
| | **MLA-2.Q2:** Do you see the involvement of a clinician beneficial in discussing how to balance the tradeoffs? **Please choose from Yes/ No/ Unsure** and provide some rationale for your answer | **MLA-2.Q2_A** - in my mind from a customer perspective no, we provide a tool to customers with a set sens/spec as per the approved CE marking i.e. this isn't then customisable. However we have in house clinicians who advise on the development of new products on what would/wouldn't be acceptable as a tradeoff | **MLA-2.Q2_A:** Yes, in particular for what performance metrics to optimised the model to. For the case of identifying patients at risk of becoming high-cost high-need patients, we want to ensure false negatives do not include any events related to mortality. Clinicians will have an holistic view that will include the healthcare system, the patient and their health<br><br>Comment: I guess the clinician will also have a view of what extreme events we can use to validate the FNs beyond mortality? |
| **MLA-3:** Several measures are available to assess some of these trade-offs. For example the Area Under ROC Curves enable the trade-offs between false-positive and false-negative classifications to be evaluated. | **MLA-3.Q1:** What would be the best guidance to provide for the types of ML performance metrics that should be taken into account for healthcare? | **MLA-3.Q1_A:** | **MLA-3.Q1_B:** It will depend on the purpose of the task. When optimising the model, we focus on AUC and false negatives because the output of the model is always screened by a clinician, thus reviewing all false positives. <br><br> **Comment:** It won't be superfluous to also describe what it will mean if one of the metrics is chose for optimisation. E.g. if you choose false negatives that means you will have higher sensitivity and lower positive predictive value. Again helping to build a common understanding on related (But often assumed similar metrics) <br><br> Perhaps worth considering that other metrics might be needed for regression based problems? |
| **MLA-4:** Rational and decisions made need to be logged in the Model Development Log | **MLA-4.Q1:** Who do you see managing the Model Development Log? | **MLA-4.Q1_A:** Applied ML Lead | **MLA-4.Q1_B: Comment:** The project/workstream owner to map out the process, where the ML engineers have to mark which stages will have the highest impact on the performance of the model and all changes made to these key pivots are to be logged |

# Model Verification Assurance (MVA)

| Key Discussion Point | Review Question | Review Answer - A | Review Answer - B |
|---|---|---|---|
| **MVA-1:** Model verification may consist of two sub-activities: test-based verification and formal verification. For every ML safety requirement at least one verification activity shall be undertaken. | **MVA-1.Q1:** Do you agree with key discussion point **MVA-1** as a sensible approach to verfication? Please choose **from Yes/ No/ Unsure** and provide a justification for your chosen option. | **MVA-1.Q1_A:** At each new site we engage with we run a pass of our current ML model. Depending on performance we will calibrate with this new site data if needed. Additionally we have run and intend to run formal clinical investigations on performance in both a double reading and standalone workflow | **MVA-1.Q1_B:** Yes, we currently verify our models with test-based verification. |
| | **MVA-1.Q2:** Are these model verification (test-based/formal verification) techniques sufficent for healthcare? Please choose **from Yes/ No/ Unsure** and provide a justification for your chosen option. | **MVA-1.Q2_A:** The verification and metrics used are the same for both a local site testing and for clinical investigations - mathematical/statistical analysis of sensitivity/specificity on both double reading and standalone performance. To confirm with ML team whether any formal verifications are used, but assume not. | **MVA-1.Q2_B:** Yes, additionally models could also be tested/verified on their bias/fairness for particular unpriviledged groups. This is particular relevant in healthcare (example, google's skin cancer detection performance is lower for darker skin) |
| **MVA-2:** It is important to ensure that the verification data is not made available to the development team since if they are to have oversight of the verification data this is likely to lead to techniques at development time which circumvent specific samples in the verification set rather than considering the problem of generalisation more widely. | **MVA-2.Q1:** In what form should the healthcare AMLAS request evidence from an organisation that their ML dev team have not had sight of the verification dataset? | **MVA-2.Q1_A:** | **MVA-2.Q1_B:** This is tricky as additional data for verification might be difficult to obtain (we are given 3 years of historical data<br>1 .we perform initial EDA to verify our approach/feature skeleton captures the local population disease profile and consumption patterns<br>2. we develop the model, adapt the feature skeleton, train and validate<br>3. deployment on live data |
| | **MVA-2.Q2:** What is your opinion on - a healthcare organisation who deploys your technology using its own dataset (possibly synthetic data) to verify the model? | **MVA-2.Q2_A:** Our data sets are all real world data from partner sites | **MVA-2.Q2_B:** In our case, we are the deployment org. We do not use syntetic data at the moment to verify our models.<br><br>Example of an adverse event: patient with continuous unplanned admissions, but ICD10 codes not part of the feature skeleton, is not being identified by the prediction model.<br>Possible verification test: at what point would the pred model identify the patient just based on their activity consumption? |
| MVA-3: [Note 34] For some safety-related properties, such as interpretability, it may be necessary to include a **human in the loop** evaluation mechanism. This may involve placing the component into the application and generating explanations for experts to evaluate & [Example 35] The DeepMind retinal diagnosis system generates a segmentation map as part of the diagnosis pipeline. This image may be shown to clinical staff to ensure that the end user is able to understand the rationale for the diagnosis. As part of a verification stage these maps may be presented to users without the associated diagnosis to ensure that the images are sufficiently interpretable. | MVA-3.Q1: Should humans (healthcare professionals) be included in the verification process? | MVA-3.Q1_A: This is planned as part of both our deployment and intended clinical investigation. For example, as part of phase 2 we are engaging with our local site clinical leads to undertake a detailed discordant case review, where they will have the opportunity to further understand the ML component's outputs. Additionally, we are also planning an arbitration reader study where clinicians will have the opportunity to understand the impact of the ML component on the arbitration routes as well as a prospective study where the component will be running on live cases under a research setting where clinicians will work with the tool in a live clinical workflow (can dig into the protocol for more detail here) | MVA-3.Q1_B: Humans with the clinical and technical knowledge could be an additional step to verify the outputs of the model or test edge cases. |

# Model Deployment Assurance (MDA)

| Key Discussion Point | Review Question | Review Answer - A | Review Answer - B |
|---|---|---|---|
| MDA-1: AMLAS states the deployment process shall be followed not only for initial deployment of the component but also for any subsequent deployment required to update the component within the system. | MDA-1.Q1: Prior to deployment should there be a distinction between ML component which is First of Type (FoT), going for Full Rollout (FR) or existing live ML components being updated through Request For Change (RFC)? | MDA-1.Q1_A: At the moment we are working through our first time rollouts for a full roll out. The ML model that is ultimately deployed will be frozen. Any updates to the model will need to go through an additional CE marking process before getting regulatory approval for use. Then we would roll out to all of our sites. We are currently exploring the method of using a change advisory board (CAB) for this. Once the tool is fully deployed on sites there will be formal KPIs, performance metrics, up times etc.. that we will be commerically contracted to deliver. As part of this contract there will also be a process implemented for roll out of upgrades. Current thinking is through a CAB, but detailed still being worked through.<br><br>Note that any model updates will just be cloud pushed in - no physical changes or work needed | MDA-1.Q1_B: Yes, should keep track of what changed and its impact on the outputs etc (test against adverse scenarios - model verification).<br><br>Also ask "why are we changing the model?"<br>- because monitoring scripts identified a problem, so we retrain/modify to avoid the problem. This might trigger another problems (adverse events), which means a new verification method would need to be worked on.<br>Example: missing data for one feature has increased significantly, so we replace it with a proxy in the new version of the model.<br><br>Important to distinguish FoT and RFC:<br>FoT - simply focys on model performance<br>RFC - model verification is more important here? |
| MDA-2: Activity 15 seems to infer the manufacturer deploys onto the intended hardware and does the safety activity. | MDA-2.Q1: What role does the deploying organisation need to play here? | MDA-2.Q1_A: Our phased deployment has vast involvement from the deploying organisation; particularly the clinicians. We have three phases of analysis done on retrospective data each of which concludes with a write up summary report of the analysis which needs to be signed off by clinicans before proceeding intot the next phase. The three types of retrospective analysis done in the three phases are; a general population test of generalisabilty, a case by case discordance review and a workflow arbitration simulation.<br>Additionally, we will be undertaking system integration testing with deployment site IT teams and PACS teams. All conducted in system test environments usually with dummy data | MDA-2.Q1_B: We are the manufacturer and deploying org.<br>Development org sets a quality criteria for the model.<br>Deploying org needs to ensure the outputs of the deployed model meet those criteria.<br><br>In summary:<br>- ensure that it works (system environment)<br>- ensure it works well (meets quality/ safety criteria)<br><br>Example:<br>If deployment environment is different from model development, deployment org needs to perform adjustments and ensure performance is maintained. |
| | MDA-2.Q2: Should the manufacturer and deploying organisation compare safety cases (still being worked on) prior to deploying into the target environment? | MDA-2.Q2_A: We are currently doing this through the DCB0129 / DCB 0160 process. We have completed the deliverables for the 129 and the deploying organisation will be completing the 160.<br>However given the nature of our work, we don't yet have national approval to move into use in a live setting, we are currently only working on retrospective analysis and parallel clinical investigatinos in the hope to generate evidence to change national stakeholder opinion. At this point the requirements of the use of AI in screening may be different.<br>For now the safety cases are being written from a holistic perspective with a view of the retrospective analysis and the current plans for live use (which may evolve as evidence is generated and national stakeholders input into this) | MDA-2.Q2_B: Yes.<br>- model performance checks (for example ensure model is fair and does not discriminate)<br>- model verification |
| MDA-3: Measures shall be put in place to monitor and check validity throughout the operation of the system of the key system and environmental assumptions. Mechanisms shall be put in place to mitigate the risk posed if any of the assumptions are violated<br><br>Deployment org would have a big role in this as it is unrealistic for a manufacturer to engineer self-monitoring which triggers a timely mitigating intervention.<br>For e.g. if system malfunctions, mitigation may need to come from a human (handover) which would most likely be a deploying org HCP. | MDA-3:Q1: Could you comment on whether healthcare organsiations you work with would have the right personnel in place to address MDA-3? | MDA-3:Q1_A: Yes, we have teams (software engineers) currently working on the integrations for online monitoring of the tool. | MDA-3:Q1_B: Since we fulfill both roles, we are the ones who would monitor and escalate with the hospital in case something goes wrong.<br>Our point of contact is usual the local informatics team.<br><br>But because it is not always expected that the manufacturer will be deploying their AI model, the healthcare org should have in place a local team who has the knowledge and capability for deployment, monitoring and risk mitigation. |
| | MDA-3:Q2: Is there a need for a certified/qualified AI-Health Care Professional or AI-Clinical Safety Officer? | MDA-3:Q2_A: Do you mean within the deployment organisation? Or within the manufacturer? Our in house clinicians have recently gone through the NHS D provided Clinical Safety Officer training. Their insights are also deeping involved in product development internally. At our deploying organisations I'm not aware of any specific AI focussed health care professionals, we are mainly dealing with Consultant Radiologists | MDA-3:Q2_B: Yes, both clinical and technical. Someone who can verify the ML on a clinical safety basis. Someone who can ensure Governance requirements are being met at the minimum. Someone who can verify the ML on a technical safety basis (AI approach/performance metrics understanding/monitoring/risk mitigation/can question/challenge). Additionally, day to day users of the ML outputs should be knowlegeable enough to spot any system malfunction. |
| MDA-4: The system shall monitor the outputs of the ML model during operation, as well as the internal states of the model, in order to identify when erroneous behaviour occurs. These erroneous outputs, and model states, shall be documented in the erroneous behaviour log ([DD]) | MDA-4.Q1: AMLAS places much emphasis on the the [DD] erroneous behaviour log. Who should fill this log in? | MDA-4.Q1_A: I would imagine this is what is being referred to in my answer to MDA3-Q1 in terms of the cloud clinical monitoring. The outputs of the model are contunually monitored for performance in terms of the accuracy of its outputs and any drop in so would trigger an alert. My expectation would be that this would fall into an automatic log. | MDA-4.Q1_B: The development organisation should create the structure of the erroneous log in partnership with the deployment organisation. However, it is the deployment's organisation responsibility to track and fill in the log. The deploying organisation should monitor the inputs for the prediction model, where if they match a certain quality criteria then the outputs of the prediction model should also meet that criteria. For example, a "classic/standard" case might be developed for the model/algorithm which should always receive a high risk score. The computation of that "classic/standard" case should be done with a different language/programme (R vs Python) than the one being used for deployment. A second algorithm might also be used to ensure that erroneous behaviour is minimised (e.g. a high-risk scoring patient in a given model such as 3 or more bed days, should also be a high-risk scoring patient for a model predicting patients at risk of 2 or more bed days) Does erroneous output mean unintended answer to a |
| MDA-5: As well as considering how the system can tolerate erroneous outputs from the ML model, integration shall consider erroneous inputs to the model. Very important for assurance - injecting erroneous inputs from adversarial behaviour or through 'work as is' vs 'work as imagined'. | MDA-5.Q1: Would you agree the healthcare organisation would need to provide expertise to address this point? | MDA-5.Q1_A: Yes 100% This is something we are currently working through and is a huge challenge. We are looking for ways that the system tags cases that shouldn't be read by the ML component. For example where is it tagged in the clinical system that a woman has an implant and therefore her mammogram shouldn't be read by the ML. This is a challenge to automate that we are working through with huge input from our site partners. For example we have been doing detailed walkthroughs of the patient pathway from arrival to screening to result receiving to understand exactly where each piece of information is logged and where it can automatically alert the ML to not read a case. Or if the ML does read this case it needs to be tagged as going through a separate pathway or to ignore the ML's output for these types of cases | MDA-5.Q1_B: Yes. The deployment organisation should consult with the development organisation to monitor the most important features/variables so that if any sigificant deviations occur a corrective action can be done or at least considered/discussed between all stakeholders. |